\documentclass[table,conference]{IEEETran}
\IEEEoverridecommandlockouts

\usepackage{amsmath}  	
\usepackage{booktabs} 	
\usepackage{caption}    
\usepackage{fancyvrb} 	
\usepackage{float} 		
\usepackage{graphicx} 	
\usepackage{hyperref}   
\usepackage{lipsum}   	
\usepackage{microtype}  
\usepackage{multicol} 	
\usepackage{multirow} 	
\usepackage{tcolorbox}
\usepackage{units}    	
\usepackage{wrapfig}
\usepackage{url}
\usepackage{highlight} 	
\usepackage{color,soul}
\usepackage{longtable}

\usepackage{xspace} 
\usepackage{makeidx} 

\usepackage{tcolorbox}
\definecolor{lightnavy}{HTML}{F2F2F9}
\definecolor{whitesmoke}{HTML}{F5F5F5}
\definecolor{whitesmoke}{HTML}{FFFFFF}
\definecolor{navyblue}{HTML}{00007B}
\definecolor{lightgreen}{HTML}{F2FBF8}
\definecolor{green}{HTML}{88D6D1}
\definecolor{darkgreen}{HTML}{2A7F7F}
\definecolor{lightpurple}{HTML}{E5E5FF}
\definecolor{darkpurple}{HTML}{3F3FA3}
\definecolor{black}{HTML}{000000}

\tcbuselibrary{theorems,skins,hooks}

\newtcbtheorem[number within=section]{Process}{Process}
{%
  enhanced
  ,colback = white
  ,boxrule = 0sp
  ,borderline west = {5pt}{0pt}{black}
  ,sharp corners
  ,detach title
  ,before upper = \tcbtitle\par\smallskip
  ,coltitle = black
  ,fonttitle = \bfseries\sffamily
  ,description font = \mdseries
  ,terminator sign dash
  ,separator sign none
}
{th}

\newtcbtheorem[number within=section]{Example}{Example}
{%
  enhanced
  ,colback = lightnavy
  ,frame hidden
  ,boxrule = 0sp
  ,borderline west = {5pt}{0pt}{navyblue}
  ,sharp corners
  ,detach title
  ,before upper = \tcbtitle\par\smallskip
  ,coltitle = navyblue
  ,fonttitle = \bfseries\sffamily
  ,description font = \mdseries
  ,terminator sign dash
  ,separator sign none
}
{th}

\newtcolorbox{Definition}[1]
{
  enhanced
  ,colback = lightpurple
  ,colframe = darkpurple
  ,coltitle = darkpurple
  ,colbacktitle = lightpurple
  ,fonttitle = \bfseries
  ,title = {#1}
  ,attach boxed title to top right = {yshift = -5pt, xshift = -7mm}
  ,boxed title style = { boxrule = .25mm }
  ,arc = 5mm
  ,interior code app =
    {
      \node
      [
        anchor=south west
        ,line width = 0.5mm
        ,rounded corners
        ,inner sep = 5pt
        ,draw = darkpurple
        ,fill = lightpurple
        ,yshift = -5pt
        ,xshift = 7mm
        ,text = darkpurple
        ,font = \bfseries
      ] at (frame.north west)
      {\ Definition\ \null};
    }
}

\newtcolorbox{BlackBox}[1]
{
  enhanced
  ,title = {#1}
  ,arc = 5mm
  ,colback = white
  ,boxrule = 0sp
  ,borderline west = {5pt}{0pt}{black}
  ,sharp corners
  ,detach title
  ,before upper = \tcbtitle\par\smallskip
  ,coltitle = black
  ,fonttitle = \bfseries\sffamily
  ,description font = \mdseries
  ,terminator sign dash
  ,separator sign none
}
\usepackage{epigraph}

\graphicspath{ {graphics/} } 				            
\renewcommand{\opacity}{50}
\hypersetup{
    colorlinks = true,
    citecolor = {blue},
    urlcolor = {blue},
}

\def\BibTeX{{\rm B\kern-.05em{\sc i\kern-.025em b}\kern-.08em
T\kern-.1667em\lower.7ex\hbox{E}\kern-.125emX}}

\begin{document}

\title{Evaluation of Habitat Robotics using Large Language Models}

\author{\IEEEauthorblockN{William Li}
    \IEEEauthorblockA{\textit{Artificial Intelligence Group} \\
        \textit{MIT Lincoln Lab}\\
        William.Li@ll.mit.edu}
    \and
    \IEEEauthorblockN{Lei Hamilton}
    \IEEEauthorblockA{\textit{Artificial Intelligence Group} \\
        \textit{MIT Lincoln Lab}\\
        Lei.Hamilton@ll.mit.edu}
    \and
    \IEEEauthorblockN{Kaise Al-natour}
    \IEEEauthorblockA{\textit{Artificial Intelligence Group} \\
        \textit{MIT Lincoln Lab}\\
        Kaise.Al-natour@ll.mit.edu}
    \and
    \IEEEauthorblockN{Sanjeev Mohindra}
    \IEEEauthorblockA{\textit{Artificial Intelligence Group} \\
        \textit{MIT Lincoln Lab}\\
        smohindra@ll.mit.edu}

}

\maketitle

\begin{abstract}
    \noindent This paper focuses on evaluating the effectiveness of Large Language Models at solving embodied robotic tasks using the Meta PARTNER benchmark.  Meta PARTNR provides simplified environments and robotic interactions within randomized indoor kitchen scenes.  Each randomized kitchen scene is given a task where two robotic agents cooperatively work together to solve the task.  We evaluated multiple frontier models on Meta PARTNER environments.  Our results indicate that reasoning models like OpenAI o3-mini outperform non-reasoning models like OpenAI GPT-4o and Llama 3 when operating in PARTNR's robotic embodied environments.  o3-mini displayed outperform across centralized, decentralized, full observability, and partial observability configurations.  This provides a promising avenue of research for embodied robotic development.
\end{abstract}

\begin{IEEEkeywords}
    Embodied Robotics, Meta PARTNR, Large Language Models
\end{IEEEkeywords}

\section{Background}

In 2020, OpenAI released GPT-3\cite{brown2020languagemodelsfewshotlearners} which ignited considerable interest from the research community regarding Large Language Models (LLM).  In particular, robotics researchers began looking toward LLMs as a way of integrating intelligence into their automata.  Prior to the popularity of LLMs, most researchers were focused on designing robotic controls or leveraging reinforcement learning\cite{mu2021maniskillgeneralizablemanipulationskill} to perform robotic manipulation.  With the release of ever more powerful LLMs, robotics researchers began looking at ways to incorporate LLMs into their development.

In 2023, a global consortium of researchers from industry and academia released the Open-X\cite{embodimentcollaboration2024openxembodimentroboticlearning} dataset for robotic research. The Open-X dataset provided a dataset which could be used to train robotic manipulation with corresponding images and natural language instructions.  This paved the way for developing Vision Language Action (VLA) models which integrated language, vision, and robotic control.

In 2024, Berkley University released  Octo\cite{octomodelteam2024octoopensourcegeneralistrobot}; a seminal paper on using VLA Transformer to perform robotic manipulation.  In that same year OpenVLA\cite{kim2024openvlaopensourcevisionlanguageactionmodel} was released, demonstrating improved VLA performance built on top of Llama2\cite{touvron2023llama2openfoundation}.  Diffusion methods such as RDT-1B\cite{liu2025rdt1bdiffusionfoundationmodel} further improved upon OpenVLA to demonstrate additional performance gains under certain situations.

In parallel to robotic research, the number of simulators supporting robotic research was growing as well.  Popular robotic simulators have emerged such as Maniskill\cite{mu2021maniskillgeneralizablemanipulationskill}, Nvidia Isaac Sim\cite{mittal2023orbit}, AI2Thor\cite{kolve2022ai2thorinteractive3denvironment}, Robocasa\cite{robocasa2024} and Genesis\cite{Genesis}.  Many of these simulators were designed to support robotic manipulation by rendering physics accurately.

In 2024, Meta released a benchmark designed specifically to study embodied robotics in collaborative environments with explicit LLM support.  The benchmark was called Planning and Reasoning in Embodied Multi-agent Tasks (PARTNR)\cite{chang2024partnrbenchmarkplanningreasoning}; it was designed to study robot coordination using Large Language Models (LLMs) for indoor scenes.  PARTNR was built on the Habitat\cite{savva2019habitatplatformembodiedai} simulator and was specifically designed to benchmark heterogeneous multiagent coordination using LLMs:

\epigraph{PARTNR tasks exhibit characteristics of everyday tasks, such as spatial, temporal, and heterogeneous agent capability constraints. We employ a semi-automated task generation pipeline using Large Language Models (LLMs), incorporating simulation in the loop for grounding and verification. PARTNR stands as the largest benchmark of its kind, comprising 100,000 natural language tasks, spanning 60 houses and 5,819 unique objects.}{PARTNR: \\A Benchmark for Planning and Reasoning \\in Embodied Multi-agent Tasks}

PARTNR provides a standardized baseline for evaluating \textit{planning}, \textit{coordination}, and \textit{perception} for multiagent embodied robots in simulated indoor settings.  The tasks require \textit{planning} because agents must apply the correct sequence of actions in order to accomplish their tasks.  PARTNR simulates multiagent collaborative tasks which require \textit{coordination}.  Environments can be setup as either fully observable or partially observable to test the \textit{perception} capabilities of LLMs.  Tasks and environments are procedurally generated ensuring a diverse range of unique tests.

Despite state-of-the art performance of LLMs in advance math\cite{liu2024mathbenchevaluatingtheoryapplication}, graduate level questions\cite{rein2023gpqagraduatelevelgoogleproofqa}, and coding capabilities\cite{deepseekai2025deepseekr1incentivizingreasoningcapability,deepseekai2025deepseekv3technicalreport,openai2024openaio1card,openai2024gpt4ocard}; LLMs lack an embodied understanding of the physical world.  This gap between LLM's understanding of the physical world prevents generalized deployment of robotics into the real world.  In the PARTNR\cite{chang2024partnrbenchmarkplanningreasoning} paper they stated that state-of-the art LLM models displayed poor performance in embodied tasks:

\epigraph{
    The analysis reveals significant limitations in SoTA models, such as poor coordination and failures in task tracking and recovery from errors\dots  Overall, PARTNR highlights significant challenges facing collaborative embodied
    agents and aims to drive research in this direction\dots}{PARTNR: \\A Benchmark for Planning and Reasoning \\in Embodied Multi-agent Tasks}

Since the release of the original PARTNR benchmark, multiple new frontier models have been released such as OpenAI GPT-4o, OpenAI o3-mini, DeepSeek R1, etc.  OpenAI GPT-4o is considered a non-reasoning model whereas DeepSeek R1 and OpenAI o3-mini are considered reasoning models.  The purpose of our study was to determine how well these models perform on the PARTNR benchmark.  Chain of Thought (CoT), used in reasoning models, has shown dramatically improved performance over non-reasoning models using test time compute.  However, they utilize significantly more tokens and often take longer to produce an answer.  We wanted to understand whether reasoning models that utilized CoT could help with embodied tasks in robotics.

\section{Simulation}

PARNTR provides a common framework for embodied robotic agents to interact within a digital indoor environment to solve a task.  Two robotic agents are simulated within the environment: (1) a humanoid robot and (2) a Boston Dynamics Spot robot.  It should be noted that the spot robot and the humanoid robot have different capabilities.  The humanoid robot can perform all the actions of the Spot robot, but has additional capabilities.

\textbf{Spot Robot Capabilities:}
\begin{itemize}
      \item \textbf{Close:}
            Used for closing an articulated entity. You must provide the name of the furniture you want to close.
      \item \textbf{Explore:}
            Search a specific room by visiting various receptacles or furniture in that room. The input to the skill is the EXACT name of the room to be visited. Use the room names provided in the house description. This tool exhaustively explores the specified room.
      \item \textbf{Navigate:}
            Used for navigating to an entity. You must provide the name of the entity you want to navigate to.
      \item \textbf{Open:}
            Used for opening an articulated entity. You must provide the name of the furniture you want to open.
      \item \textbf{Pick:}
            Used for picking up an object. You must provide the name of the object to be picked. The agent cannot hold more than one object at a time.
      \item \textbf{Place:}
            Used for placing an object on a target location. You need to provide the name of the object to be placed, the name of the furniture where it should be placed, spatial relation ("on" or "within") describing the relation between the object and furniture. The object to be placed must already be held by the agent (i.e. picked previously).
      \item \textbf{Rearrange:}
            Used for moving an object from its current location to the target location. You need to provide the name of the object to be moved, the name of the furniture where it should be moved, spatial relation ("on" or "within") describing the relation between the object and furniture. This will automatically pick the specified object and move to the target furniture and attempt to place it.
      \item \textbf{Wait:}
            Used to make agent stay idle for some time.
\end{itemize}

\textbf{Humanoid Robot Capabilities:} The humanoid robot has all the capabilities of the Spot robot and a few additional capabilities:

\begin{itemize}
      \item \textbf{Clean:}
            Used for cleaning an object. You need to provide the name of the object to clean.
      \item \textbf{Fill:}
            Used for filling an object. You need to provide the name of the object to fill.
      \item \textbf{Pour:}
            Used for pouring from one container to another. This skill will pour into the specified container from whichever container is currently held by the agent.
      \item \textbf{PowerOff:}
            Used for turning off a powered object. You need to provide the name of the object to be turned off.
      \item \textbf{PowerOn:}
            Used for turning on a powered object. You need to provide the name of the object to be turned on.
\end{itemize}

When designing our experiments we focused on two key areas: (1) \textit{centralized planning} versus \textit{decentralized planning} and (2) \textit{fully observable} versus \textit{partially observable}.  We wanted to understand how well frontier LLMs could correctly plan out their actions in randomized indoor scenes with different levels of observability.  We were not necessarily concerned about the physical interactions (i.e. motor controls) of the robots, thus we abstracted away motor controls.  A robot could perform an action on an object as long as it was close to the object.

\begin{figure*}
      \includegraphics[width=0.85\paperwidth]{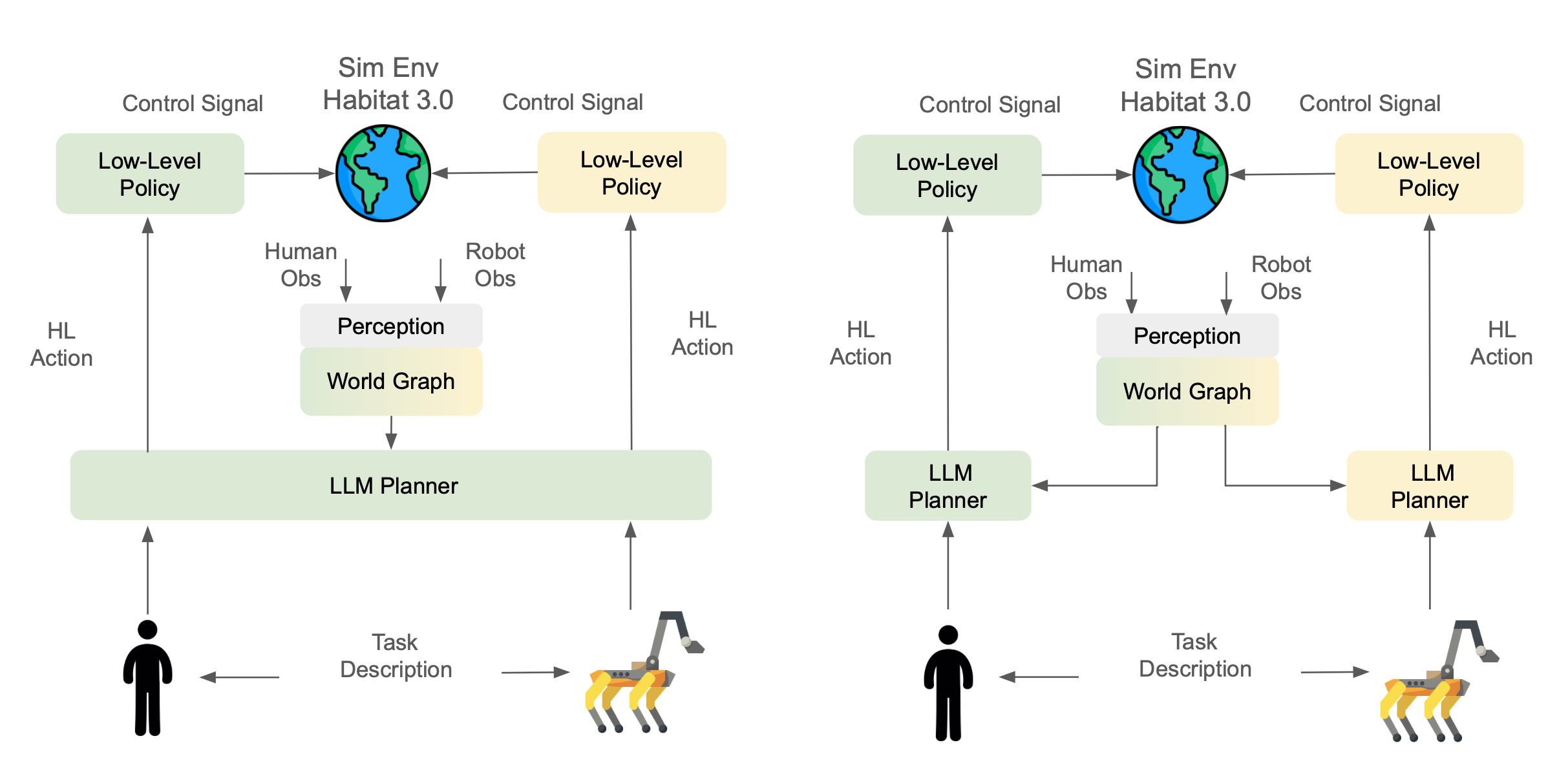}
      \caption{\textbf{Centralized vs Decentralized Planning:} In the centralized planner, one planner controls the actions of both agents.  The actions are sent to a low-level policy which mechanically performs the actions.  These actions are then updated within the simulator (Habitat 3.0) and outputs a new state which is then observed individually by each agent.  The observations update a shared world graph which is fed back into the centralized planner repeating the process.  The decentralized planner in contrast has separate planners for each agent.  In decentralized planning the agents have individual observations which are used to update a non-shared world graph.  The figure was sourced from PARTNR\cite{chang2024partnrbenchmarkplanningreasoning}.}
      \label{fig:central_vs_decentral}
\end{figure*}

\begin{itemize}
      \item \textbf{Centralized vs Decentralized Planning:}
            In centralized planning, a single LLM decides actions for both agents, with complete information about both agent's states; effectively removing any need for coordination between the agents. In decentralized planning, each agent is controlled by a different LLM, and each LLM needs to reason about the other agent's actions. See Figure~\ref{fig:central_vs_decentral}.
      \item \textbf{Partial vs Full Observability:}
            In partially observed environments the planner knows the house's layout, but objects are hidden.  The location of hidden objects can only be discovered using exploration actions. In fully observed environments room layout and all object locations are visible.
\end{itemize}

Iterating over all possible combinations, we have Table-\ref{tab:experiment_design}:

\begin{table}[ht]
      \centering
      \fontfamily{ppl}\selectfont
      \begin{tabular}{lll}
            \toprule
            Planner       & Observability & Difficulty \\
            \midrule
            Centralized   & Full          & Easy       \\
            Centralized   & Partial       & Moderate   \\
            Decentralized & Full          & Moderate   \\
            Decentralized & Partial       & Hard       \\
            \bottomrule
      \end{tabular}
      \caption{Design matrix for setting up experiments.}
      \label{tab:experiment_design}
\end{table}

Centralized planning is considered an easier task than decentralized planning.  Full observability is considered easier than partial observability.  For each experimental design configuration, we provide a difficulty estimate based on the combination of different parameters.  From an information standpoint, the decentralized planner can be thought of partial observability because each planner must perform planning without knowing what the other planner is doing.  The difficult assessment is really a ranking of how much information the planner has to perform its planning cycle given the limited information it has access to.

PARTNR provides a way for LLMs to interact with its environment purely using text.  This is an oversimplification because real robotics would use a live camera feed for scene perception.  For the purposes of our study, we were focused on planning and coordination of frontier LLMs in embodied robotics; camera perception was not our focus, so we decided to use a text based interactive environment.

At the start of an episode, the agents would be randomly placed somewhere in the environment.  Rooms and objects are randomized.  PARTNR would then generate a task in natural language for the indoor simulations.  Consider Example~\ref{ex:example_task} where the environment is partially observable, the agent is tasked with finding the sushi mat and move it to the living room table.  The sushi mat is not directly observed meaning the agent must explore the indoor environment to find the sushi mat.  After finding the sushi mat, it must navigate to it, pick it up, navigate to the living room table, and then place it on the living room table.

\begin{Example}{Example Task \& State of Knowledge}{Partially Observable}

      \textbf{Task:} Move the \textbf{\textcolor{brown}{sushi mat}} from the \textbf{\textcolor{red}{kitchen counter}} to the  \textbf{\textcolor{blue}{living room table}}.

      \medskip
      \textbf{State Information:}
      \begin{itemize}
            \item \textbf{living\_room\_1:}
                  \begin{itemize}
                        \item floor\_living\_room\_1
                        \item \textbf{\textcolor{blue}{table\_14}}
                        \item \textbf{\textcolor{blue}{table\_15}}
                        \item couch\_23
                        \item chair\_32
                  \end{itemize}
            \item \textbf{bathroom\_1:}
                  \begin{itemize}
                        \item floor\_bathroom\_1
                        \item toilet\_30
                  \end{itemize}
            \item \textbf{kitchen\_1:}
                  \begin{itemize}
                        \item stool\_18
                        \item stool\_19
                        \item \textbf{\textcolor{red}{counter\_34}}
                        \item cabinet\_35
                        \item cabinet\_41
                  \end{itemize}
            \item \textbf{dining\_room\_1:}
                  \begin{itemize}
                        \item floor\_dining\_room\_1
                        \item chair\_10
                        \item chair\_11
                        \item chair\_12
                        \item chair\_13
                        \item table\_21
                  \end{itemize}
      \end{itemize}
      \medskip

\end{Example}\label{ex:example_task}

\begin{Example}{Example Plan}{Partially Observable}
      \textbf{Example Planning:}
      \begin{enumerate}
            \item Explore kitchen
            \item Navigate to kitchen counter\_34
            \item Pickup sushi\_mat
            \item Navigate to living\_room\_1
            \item Put sushi\_mat on table\_14
      \end{enumerate}
\end{Example}\label{ex:example_plan}
\section{Experiment Design}

For each test point, we ran the models for ten hours to gather statistics.  The number of episodes completed by each model in ten hours differs because reasoning models use more test time compute compared to non-reason models.  This will naturally lead to less completed episodes for reasoning models like o3-mini.  However, we provide a one sigma standard deviation for the test points, so readers can assess how much variability is around the mean.

We track four different variables for evaluating agents: \textit{simulation steps}, \textit{success rate}, \textit{percent complete}, and \textit{planning cycles}.  Of the four metrics, the percent complete is the most important as it indicates the portion of episodes it completed correctly.  The other support metrics provide additional quantifier which can be helpful in assessing the performance of the agents.

\begin{itemize}
      \item \textbf{Sim Steps:}
            The number of simulation steps until the end of an episode.  Even though there is no time limit to complete a task, agents that are able to achieve success in a shorter amount of steps is considered superior.
      \item \textbf{Success Rate:}
            Within an episode, the agents will be given multiple decision points.  During these decision points the agent must select an action to perform.  This metric is calculated as the number of successfully completed decisions divide by the total number of decision points.
      \item \textbf{Percent Complete:}
            This metric indicates the fraction of episodes where the agent successfully completed the task.  It is calculated as the fraction of successful episodes divided by total non-crashed episodes.
      \item \textbf{Planning Cycles:}
            Planning cycles is how often the agent needs to replan its actions within an episode.  Less replanning indicates that the agent is better at long-term planning.
\end{itemize}

\section{Results}

\begin{table*}[h]
    \centering
    \begin{tabular}{lllrrrr}
        \toprule
        Model              & Planner       & Observability & Sim Steps ↓ & Success Rate ↑      & Percent Complete ↑ & Planning Cycles ↓
        \\
        \midrule
        Llama 3 8B         & Centralized   & Full          & 1338 ± 579  & 0.38± 0.5           & 56 \%              & 15.19 ± 5.8       \\
        Llama 3 8B         & Centralized   & Partial       & 2009 ± 2197 & 0.3 ± 0.5           & 46 \%              & 17.28 ± 9.5       \\
        Llama 3 8B         & Decentralized & Full          & 2856 ± 2724 & 0.37 ± 0.5          & 65 \%              & 18.08 ± 11        \\
        Llama 3 8B         & Decentralized & Partial       & 3400 ± 2910 & 0.41 ± 0.5          & 66 \%              & 16.55 ± 10.4      \\
        \midrule
        Llama 3.1 70B      & Centralized   & Full          & 1347 ± 33   & 0.74 ± 0.01         & 88 \%              & 17 ± 0.3          \\
        Llama 3.1 70B      & Centralized   & Partial       & 2298 ± 61   & 0.74 ± 0.01         & 85 \%              & 20 ± 0.5          \\
        Llama 3.1 70B      & Decentralized & Full          & -           & -                   & -                  & -                 \\
        Llama 3.1 70B      & Decentralized & Partial       & 3295 ± 76   & 0.73 ± 0.01         & 86 \%              & 15 ± 0.3          \\
        \midrule
        Llama 3.0 70B      & Centralized   & Full          & 1740 ± 1333 & 0.72 ± 0.4          & 84 \%              & 16.62 ± 7.6       \\
        Llama 3.0 70B      & Centralized   & Partial       & 2378 ± 1148 & 0.75 ± 0.4          & 84 \%              & 18.75±7.5         \\
        Llama 3.0 70B      & Decentralized & Full          & 2764 ± 2381 & 0.67 ± 0.5          & 84 \%              & 15.27 ± 8.2       \\
        Llama 3.0 70B      & Decentralized & Partial       & 4001 ± 3007 & 0.71 ± 0.5          & 82 \%              & 16.85 ± 9.3       \\
        \midrule
        GPT-4o             & Centralized   & Full          & 1144 ± 1477 & 0.69 ± 0.5          & 82 \%              & 11 ± 14.2         \\
        GPT-4o             & Centralized   & Partial       & 2036 ± 1192 & 0.72 ± 0.5          & 85 \%              & 13 ± 15.9         \\
        GPT-4o             & Decentralized & Full          & 1831 ± 2046 & 0.77 ± 0.4          & 88 \%              & 9 ± 8.8           \\
        GPT-4o             & Decentralized & Partial       & 5375 ± 4105 & 0.73 ± 0.4          & 86 \%              & 11 ± 10.1         \\
        \midrule
        o3-mini            & Centralized   & Full          & 2347 ± 1361 & \textbf{0.81 ± 0.4} & \textbf{89 \%}     & 16 ± 11.3         \\
        o3-mini            & Centralized   & Partial       & 2815 ± 2557 & \textbf{0.77 ± 0.4} & \textbf{86 \%}     & 16 ± 17.6         \\
        o3-mini            & Decentralized & Full          & 1952 ± 1524 & \textbf{0.78 ± 0.4} & \textbf{89 \%}     & 8 ± 7.3           \\
        o3-mini            & Decentralized & Partial       & 4977 ± 2534 & \textbf{0.77 ± 0.4} & \textbf{92 \%}     & 9 ± 5.9           \\
        \midrule
        Deepseek-Llama-70B & Centralized   & Full          & 1711 ± 2991 & 0.56 ± 0.5          & 72 \%              & 27 ± 30.4         \\
        Deepseek-Llama-70B & Centralized   & Partial       & 2397 ± 1707 & 0.56 ± 0.5          & 70 \%              & 23 ± 21.8         \\
        Deepseek-Llama-70B & Decentralized & Full          & 2112 ± 1956 & 0.71 ± 0.5          & 84 \%              & 15 ± 10.6         \\
        Deepseek-Llama-70B & Decentralized & Partial       & 4293 ± 4304 & 0.65 ± 0.5          & 82 \%              & 15 ± 10.5         \\
        \bottomrule
    \end{tabular}
    \caption{Bolded numbers indicate best.}
    \label{tab:results}
\end{table*}

For each test case, we list the model under test, the planner type, the observability, and the metrics.  We include a one sigma standard deviation for key metrics to show the variability.  See Table~\ref{tab:results} for our results.

Non-reasoning models such as GPT-4o and Llama 3 try to one-shot the problem and proceed.  If the plan fails, the models need to replan.  Non-reasoning models exhibit a lower percentage of completing their task, but they appear to complete them faster as shown in their sim steps.  If we think about why this is, we know that non-reasoning models are faster and iterate more quickly.  They are able to react to the simulation at a faster rate even if their actions have a lower success rate.

When we compare the non-reasoning models against a reasoning model such as o3-mini we see some interesting results.  O3-mini has higher episode percentage completion across the board.  We also see that each action taken by o3-mini has a higher success rate when compared to GPT-4o or Llama 3.1.  Unfortunately, o3-mini takes longer to make a decision which results in longer sim steps.  The trade-off of using reasoning models like o3-mini is that we pay a higher cost in sim steps for a better success rate and higher percentage of episode completion.

Based on these preliminary results, we would expect that reasoning models should outperform if time is a non-factor.  Reasoning models double-check their work and work out multiple solutions before choosing which one to execute.  This is analogous to Daniel Kanhneman's\cite{Kahneman:2011fj} System 1 versus System 2 thinking.  System 1 is reactive and immediate whereas System 2 must consider multiple possibilities and evaluate each possibility individually.

When we benchmarked DeepSeek-Llama-70B we saw that it underperformed Llama 3 70B.  This would contradict our earlier assessment that reasoning models are superior.  However, DeepSeek-Llama-70B is a base Llama 70B model which has been distilled from DeepSeek R1.  DeepSeek-Llama-70B does not have Multi-Head Latent Attention\cite{deepseekai2024deepseekv2strongeconomicalefficient} or Mixture of Experts\cite{dai2024deepseekmoeultimateexpertspecialization} present in DeepSeek R1.  While distillation is helpful in structuring a model's output to follow a CoT process, DeepSeek-Llama-70B should be viewed as a deprecated and miniaturized DeepSeek R1 which is not representative of frontier reasoning models.

We provide an example~\ref{ex:deepseek-planning} of DeepSeek-Llama-70B's reasoning to see how an agent can think about a problem.  We can see from DeepSeek-Llama-70B's internal thought process, it correctly breaks down the problem step by step.  It realizes that the sushi mat is hidden and that it needs to find where it is.  The model also identifies the correct action syntax "EXPLORE" that needs to be performed in the kitchen in order to find the sushi mat.

The LLMs under test would often have an incorrect spatial understanding of where they are relative to the object they want to interact with as shown in example~\ref{ex:failed_pickup} (red text).  In our example, the simulator outputs a response to the agent stating that it is not close enough to the object.  The LLM is able to recover from the error and fix the problem.  This type of spatial misunderstanding was prevalent across every LLM we tested.

The o3-mini model does not output its thought process, thus we were not able to identify errors in its planning.  But based on the actions we observed from o3-mini it was clear that it struggled with spatial understanding.  The simulator would often provide a helpful text response that allowed the LLMs to recover and continue.  In general, the more complex the spatial understanding needed to solve a task, the worse the LLMs would perform.

\begin{Example}{Example: DeepSeek Llama 70B - Thought Process}{example-deepseek-planning}

    \textbf{Task:} Move the sushi mat from the kitchen counter to the living room table. \\

    \textbf{Thought:} Alright, so I'm trying to figure out how to move the sushi mat from the kitchen counter to the living room table. Let's break this down step by step. First, I know that the sushi mat is an object, and according to the problem, I haven't found any objects yet. That means I need to start by locating the sushi mat. \\

    Looking at the furniture list, the sushi mat is likely to be on the kitchen counter, which is counter\_34 in kitchen\_1. So, my first thought is that I need to explore the kitchen to find the sushi mat. The Explore action is used to search a specific room thoroughly, so that makes sense here. By exploring kitchen\_1, I can check all the relevant spots where the sushi mat might be. \\

    I remember that each room doesn't need to be explored more than once, so if I don't find the sushi mat in the kitchen, I'll move on to the next room. But since the task specifies that the sushi mat is on the kitchen counter, I can focus my efforts there first. \\

    Once I find the sushi mat, I'll need to pick it up using the Pick action. Then, I'll navigate to the living room table, which is either table\_14 or table\_15. After that, I'll place the sushi mat on the table using the Place action. \\

    But before all that, the first logical step is to Explore[kitchen\_1] to locate the sushi mat...

\end{Example}\label{ex:deepseek-planning}

\begin{Example}{Example: DeepSeek Llama 70B - Spatial Confusion}{example-pickup-fail}

    \textbf{Thought:} The sushi mat has been found on the kitchen counter. I need to pick it up to move it to the living room table. \\
    \textbf{Action:} Pick[sushi\_mat\_0] \\

    \textbf{Result:} \textcolor{red}{Unexpected failure! - Failed to pick! Not close enough to the object.} \\
    \textbf{Objects:} multiport\_hub\_3: shelves\_24 in closet\_1 \\
    glass\_2: shelves\_25 in closet\_1 \\
    sushi\_mat\_0: counter\_34 in kitchen\_1 \\
    bottle\_1: shelves\_25 in closet \\

    \textbf{Thought:} I wasn't close enough to pick up the sushi mat. I need to navigate to the kitchen counter first. \\
    \textbf{Action:} Navigate[counter\_34]

\end{Example}\label{ex:failed_pickup}

\section{Discussion}

When running the experiments, we identified several problems areas where LLMs struggled when given a generated task.  These problems can be categorized as: order of operations, spatial understanding, syntax, suboptimal planning, partial observability, and early stopping.  While these categories do not encompass every possible failure case, they cover the majority of the issues we encountered.  A full description of each is provided below:

\begin{itemize}
      \item \textbf{Order of Operations:}
            Agents would often get confused regarding spatial understanding of objects.  In general, agents would need to navigate to an object in order to interact with it.  We observed agents trying to pick up objects in different rooms which they had not navigated to.  Sometimes agents would try to move, place, or rearrange an object it was not holding.
      \item \textbf{Spatial understanding:}
            At times the agents would appear to be confused as to where they were located in the environment.  This confusion led to behaviors where agents would try to interact with items in another room which is not possible.
      \item \textbf{Syntax:}
            Agents would often get confused by the syntax needed to perform an operation.  Most of the LLMs we tested were trained to code on popular programming languages prevalent on the internet.  However, PARTNR required its own custom syntax when operating robotic agents in simulated environments.  Even though the syntax was provided to the LLMs in the prompt, it would often get confused how to properly use it.
      \item \textbf{Suboptimal Planning:}
            Agents would often plan suboptimally, meaning they could accomplish the task much faster if they correctly utilized both robotic agents with balanced loads.  Often the actions assigned were unbalanced meaning one robotic agents had tasks to perform while the other agent was idle.
      \item \textbf{Partial Observability:}
            Agents would get confused by tasks that required finding hidden objects in the environment.  For example a task that required moving the sushi mat from the kitchen counter to the living room table would require first finding the sushi mat.  LLMs would sometimes try interacting with the sushi mat without realizing it needed to explore the kitchen to discover the location of the sushi mat first.
      \item \textbf{Early Stopping:}
            Agents would sometimes output "DONE" before all tasks have been complete.  This would result in a failed episode because the task was incomplete when the agent stopped the episode.
\end{itemize}

We noticed that larger LLMs generally outperformed smaller LLMs.  This was evident from the smallest model we tested (Llama 3 8B), which underperformed in all test cases.  It is estimated that The o3-mini model is closed source, so we have no insight regarding its parameter count.  However, a source from HuggingFace\cite{huggingface2025o3mini} estimates that o3-mini has 200B parameters.

For future work, it would be interesting to test with state-of-the art open-source models such as Alibaba QwQ 32B or DeepSeek R1 670B.  Alibaba's reasoning models are significantly smaller and can be run on commercial desktop Graphics Processing Units (GPU) whereas DeepSeek R1 provides frontier network architectures that are both performant and more efficient.
\section{Conclusion}

Our experiments showed that reasoning models such as o3-mini outperformed non-reasoning models like Llama 3 and GPT-4o.  While there was quite a bit of variation among the test points, o3-mini slightly outperformed the non-reasoning models in all test cases.  The out performance is statistically significant but the degree of outperform is small in each test case.

We identified and categorized multiple areas where frontier models struggled.  The problematic areas include: order of operations, spatial understanding, syntax, suboptimal planning, partial observability, and early stopping.  While humans sometimes struggle with these problems on an initial play through of an episode, humans quickly adapt and reach a near optimal solution.  LLMs often struggled with embodied robotic problems.  This is not surprising because most LLMs were trained on text scraped from the internet.  Most internet text does not require an embodied understanding grounded in physical reality.  PARTNR allows us to measure the knowledge gap of LLMs understanding of embodied indoor scenes.

Further questions that remain unresolved are whether more advance frontier reasoning models could improve performance.  Future options for research include testing Alibaba QwQ 32B or DeepSeek R1 which are considered state-of-the art open source reasoning models.  Google Gemini robotics\cite{geminiroboticsteam2025geminiroboticsbringingai} has started developing LLMs that have a better grounded understanding in embodied robotics.  This could be a promising avenue for research for robotics and LLMs.  In our testing, we only used text input LLMs, meaning we did not feed images or video to the LLMs.  Future research vectors can look at feeding image or video to VLA models.

In conclusion, leveraging LLMs for embodied robotics is difficult because models trained on text scraped from the internet do not directly translate into an understanding of physical spatial reality.  We believe that providing an embodied understanding of the physical world to LLMs would require a different type of training data.  There has already been some work in building datasets for embodied robotics such as the Open-X\cite{embodimentcollaboration2024openxembodimentroboticlearning} dataset.  This has led to new models like Google Gemini robotics.  Currently, Google Gemini has not released their models to the public, but it could be an interesting research direction for evaluation in the future.
\section{Acknowledgements}

Research was sponsored by the Department of the Air Force Artificial Intelligence Accelerator and was accomplished under Cooperative Agreement Number FA8750-19-2-1000. The views and conclusions contained in this document are those of the authors and should not be interpreted as representing the official policies, either expressed or implied, of the Department of the Air Force or the U.S. Government. The U.S. Government is authorized to reproduce and distribute reprints for Government purposes notwithstanding any copyright notation herein.

\newpage
\bibliography{main}
\bibliographystyle{ieeetr}

\end{document}